\documentclass[10pt,journal,compsoc]{IEEEtran}
\usepackage{times}
\usepackage{epsfig}
\usepackage{graphicx}
\usepackage{amsmath}
\usepackage{amssymb}
\usepackage{multirow}
\usepackage{array}
\usepackage{bm}
\usepackage{url}
\ifCLASSOPTIONcompsoc
\usepackage[nocompress]{cite}
\else
\usepackage{cite}
\fi
\usepackage{color}

\begin{document}
	
	\title{Reconstructive Sequence-Graph Network for Video Summarization}

	\author{Bin~Zhao,
		Haopeng~Li,
		Xiaoqiang~Lu,~\IEEEmembership{Senior~Member,~IEEE,}	Xuelong~Li*,~\IEEEmembership{Fellow,~IEEE}% <-this % stops a space
		\IEEEcompsocitemizethanks{
			\IEEEcompsocthanksitem Bin Zhao, Haopeng Li, Xiaoqiang Lu and Xuelong Li are with the Northwestern Polytechnical University, Xi'an 710072, China.
			(binzhao\_optimal@mail.nwpu.edu.cn)}% <-this % stops a space
		%\thanks{Manuscript received April 19, 2005; revised August 26, 2015.}
	}
	
	% The paper headers
	%\markboth{Journal of \LaTeX\ Class Files,~Vol.~xx, No.~xx, September~2019}%
	%{Shell \MakeLowercase{\textit{et al.}}: Bare Advanced Demo of IEEEtran.cls for IEEE Computer Society Journals}
	
	\IEEEtitleabstractindextext{%
		\begin{abstract}
			Exploiting the inner-shot and inter-shot dependencies is essential for key-shot based video summarization. Current approaches mainly devote to modeling the video as a frame sequence by recurrent neural networks. However, one potential limitation of the sequence models is that they focus on capturing local neighborhood dependencies while the high-order dependencies in long distance are not fully exploited. In general, the frames in each shot record a certain activity and vary smoothly over time, but the multi-hop relationships occur frequently among shots. In this case, both the local and global dependencies are important for understanding the video content. Motivated by this point, we propose a Reconstructive Sequence-Graph Network (RSGN) to encode the frames and shots as sequence and graph hierarchically, where the frame-level dependencies are encoded by Long Short-Term Memory (LSTM), and the shot-level dependencies are captured by the Graph Convolutional Network (GCN). Then, the videos are summarized by exploiting both the local and global dependencies among shots. Besides, a reconstructor is developed to reward the summary generator, so that the generator can be optimized in an unsupervised manner, which can avert the lack of annotated data in video summarization. Furthermore, under the guidance of reconstruction loss, the predicted summary can better preserve the main video content and shot-level dependencies. Practically, the experimental results on three popular datasets (\emph{i.e.}, SumMe, TVsum and VTW) have demonstrated the superiority of our proposed approach to the summarization task.
			
		\end{abstract}
		
		% Note that keywords are not normally used for peerreview papers.
		\begin{IEEEkeywords}
			key-shot, video summarization, video reconstructor, summary generator
	\end{IEEEkeywords}}
	
	\maketitle
	\IEEEdisplaynontitleabstractindextext
	\IEEEpeerreviewmaketitle
	\section{Introduction}
	\label{sec:introduction}
	
	Nowadays, huge amounts of video data are captured and shared online as the result of the ubiquitous cameras \cite{li2017multiview,DBLP:conf/eccv/ZhangGS18}. There is an imperative requirement for automatic tools to handle the large-scale video data, so that the viewers can browse them efficiently. Video summarization is the typical task raised under this background\cite{DBLP:conf/mm/ZhaoLL17,DBLP:conf/cvpr/ZhaoLL18}.
	
	Video summarization condenses the original video into a short summary meanwhile preserves the main content. There are several summary formats, including storyboard \cite{DBLP:journals/prl/AvilaLLA11}, skimming \cite{DBLP:conf/eccv/GygliGRG14}, and synopsis \cite{DBLP:journals/tip/LiWL16}. In this paper, we focus on the most popular one, \emph{i.e.}, video skimming, which is formed by several key-shots. It provides a viewer-friendly way for video browsing and can be used for many subsequent video analysis tasks \cite{hussain2020comprehensive,li2017general}.
	
	\begin{figure}
		\centering
		\includegraphics[width=0.45\textwidth]{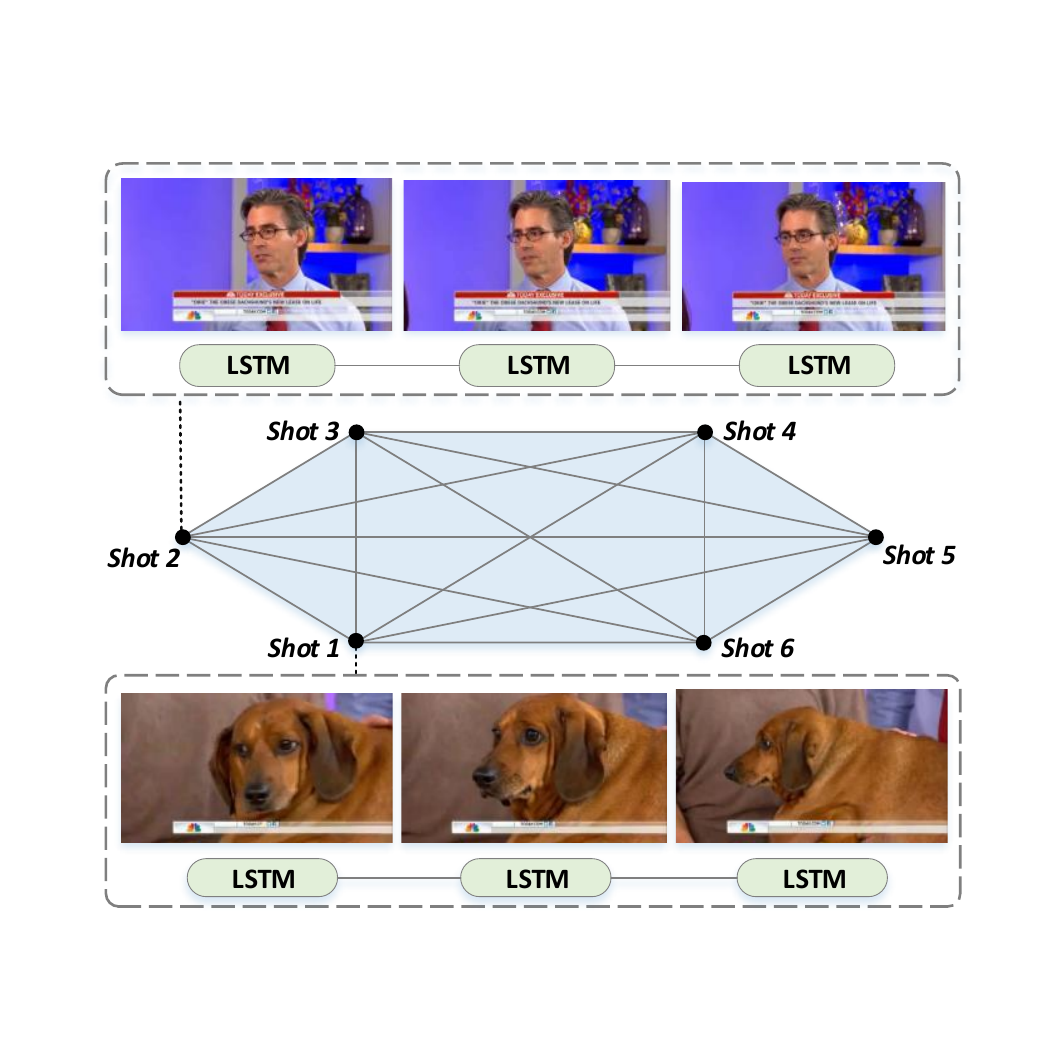}
		\caption{The sequence-graph model for video summarization. Two shots are displayed as examples.	The frames in each shot are taken as sequences and encoded by LSTM. The shots are modeled as a complete graph, where all pairwise dependencies are captured.}
		\label{fig1}
	\end{figure}
	
	Lots of video summarization approaches have been proposed in the literature. Among them, RNN-based approaches have gained significant attention recently \cite{DBLP:conf/aaai/ZhouQX18,DBLP:conf/eccv/ZhangGS18,DBLP:conf/mm/ZhaoLL17,DBLP:conf/cvpr/MahasseniLT17}. In general, these approaches cast the video data as a sequence of frames, and summarize the video by exploiting the temporal dependencies lying in the sequence. Although RNN-based approaches have made tremendous progress, these sequence models have inherent drawbacks. Specifically, they are capable of capturing the local neighborhood dependencies, but usually fail to handle the global long-distance ones and are easily distracted by noise \cite{DBLP:journals/corr/abs-1811-01824,DBLP:conf/emnlp/JiaL17}.
	
	\subsection{Motivation and Overview}
	
	Intrinsically, the video data is layered as frames and shots \cite{DBLP:conf/cvpr/PanXYWZ16}. Shot is the intermediate state between frame and video, which is formed by several consecutive frames and usually contains a certain activity. The frames in the shot are suitable to be modeled as a temporal sequence by RNN, since they are short and vary smoothly over time \cite{DBLP:conf/iccv/VenugopalanRDMD15}. But the information flow varies largely among shots, and the relationship between adjacent shots is not as tight as frames. Especially for those edited videos \cite{li2017general}, there are even no obvious temporal dependencies between the activities recorded by adjacent shots. In this case, current sequence models that just consider the neighborhood dependencies may lead to inevitable interference for the video summarization task.
	
	Actually, the sequence is a special case of graph, where only the consecutive items are connected. To better capture both the local and global dependencies among shots, it is more proper to model the video shots as a complete graph. In this case, all shots are connected as the graph nodes, and the dependencies are calculated by the interactions between two shots. Thus, the interference caused by the positional distance is averted. Inspired by this point, we propose a sequence-graph model to perform video summarization, as depicted in Figure \ref{fig1}. Specifically, RNN is utilized to capture the frame-level temporal dependencies in each shot, and the graph model is employed to capture the dependencies among different shots. Finally, the video summary is selected based on all pairwise dependencies among shots.
	
	Besides, the summary content and dependencies among key-shots are essential for the viewer to infer the original video content. Motivated by this point, a summary version of the graph model is developed as a video reconstructor, and concatenated after the sequence-graph summary generator. By adopting the reinforcement scheme for optimization, the video reconstructor can provide reward to the summary generator, which enables the generator to preserve the video content and shot-level dependencies in the summarization process. Furthermore, under the guidance of the reconstructor, the summary generator can be optimized unsupervisedly.
	
	\subsection{Contributions}
	
	The contributions of the proposed approach to video summarization lie in three folds:
	
	1) A sequence-graph architecture is designed to capture the inner-shot temporal dependencies and inter-shot pairwise dependencies with LSTM and GCN hierarchically. It can effectively avert the interference caused by the positional distance of shots.
	
	2)	A summary graph is constructed as the reconstructor to optimize the generator in an unsupervised manner, and preserve the video content and shot-level dependencies.
	
	3)	Experimental results on three benchmark datasets, including SumMe \cite{DBLP:conf/cvpr/LeeGG12}, TVsum \cite{song2015tvsum} and VTW \cite{DBLP:conf/eccv/ZengCNS16}, have demonstrated that the unsupervised version of the proposed approach is comparable with existing supervised ones, and the supervised version can achieve the state-of-the-arts in most occasions.

	\subsection{Organization}
	
	The rest of this paper is organized as follows. The related works are reviewed in Section \ref{relatedwork}. The details of the proposed approach are depicted in Section \ref{approach}. The experimental results and analysis are provided in Section \ref{experiment}. Finally, the conclusion and insight are drawn in Section \ref{conclusion}.
	\section{Related Works}
	\label{relatedwork}
	%There has been a steady development of video summarization in the literature. Existing approaches are simply classified into supervised ones and unsupervised ones. They are discussed separately in this section.
	
	\subsection{Unsupervised Video Summarization}
	
	Unsupervised approaches tackle video summarization as a subset selection problem, and focus on designing heuristic or learning-based criteria for summary generation \cite{DBLP:conf/icmcs/MeiGWHHF14,DBLP:journals/tmm/CongYL12,DBLP:conf/aaai/ZhouQX18,shemer2019ilssumm}. Clustering algorithms are widely adopted to measure the representativeness of frames or shots \cite{DBLP:journals/prl/AvilaLLA11,DBLP:conf/icip/ZhuangRHM98}. The video frames are aggregated into clusters, and cluster centers are selected as the key-frames or key-shots. Furthermore, Ngo \emph{et al.} \cite{DBLP:conf/iccv/NgoMZ03} perform clustering by modeling the video as an undirected graph. The better performance indicates the effectiveness of graph model in video summarization. Recently, Chu \emph{et al.} \cite{DBLP:conf/cvpr/ChuSJ15} propose a co-clustering algorithm to jointly summarize videos with the same topic.
	
	Besides, dictionary learning is also a widely used technique for unsupervised video summarization \cite{DBLP:conf/icmcs/MeiGWHHF14,mei2020patch,DBLP:conf/cvpr/ElhamifarSV12,DBLP:conf/cvpr/ZhaoX14a}. They summarize videos under the hypothesis that the original video content can be spanned by the elements in the summary. In \cite{DBLP:conf/cvpr/ElhamifarSV12}, the summary is modeled as a sparse dictionary to reconstruct the video. To improve the efficiency, the video is summarized in quasi-real time by learning the dictionary on-the-fly \cite{DBLP:conf/cvpr/ZhaoX14a}. %Furthermore, in \cite{DBLP:journals/tmm/LuWMGF14,DBLP:conf/icmcs/MeiGWHHF14}, the regularization terms are added to the dictionary model by taking the summary properties into consideration.
	
	\subsection{Supervised Video Summarization}
	Supervised approaches summarize the video by taking human-created summaries as references \cite{ji2020deep,xiao2020query,DBLP:conf/cvpr/SharghiLG17,khan2020blockchain,bilkhu2019attention}. Earlier works design various score functions to measure the properties that a summary is expected to hold. Naturally, frames or shots with high scores are selected into the summary. Lee \emph{et al.} \cite{DBLP:conf/cvpr/LeeGG12} and Gygli \emph{et al.} \cite{DBLP:conf/eccv/GygliGRG14} present summary property models by taking advantages of semantic information in the shot. As a result, those shots containing important and interesting objects are selected. Gong \emph{et al.} \cite{DBLP:conf/nips/GongCGS14} propose a sequential determinantal point process model to constrain the selected key-shots to be diverse with each other. Similar schemes are also adopted in \cite{DBLP:conf/cvpr/ZhangCSG16}. Lu \emph{et al.} \cite{lu2013story} conduct a stroyness model to preserve the storyline while summarization. Furthermore, Gygli \emph{et al.} \cite{gygli2015video} and Zhao \emph{et al.} \cite{DBLP:journals/tip/LiZL17} construct a comprehensive score function by combining several property models, including importance, representativeness, diversity and so on.
	
		\begin{figure*}
		\centering
		\includegraphics[width=\textwidth]{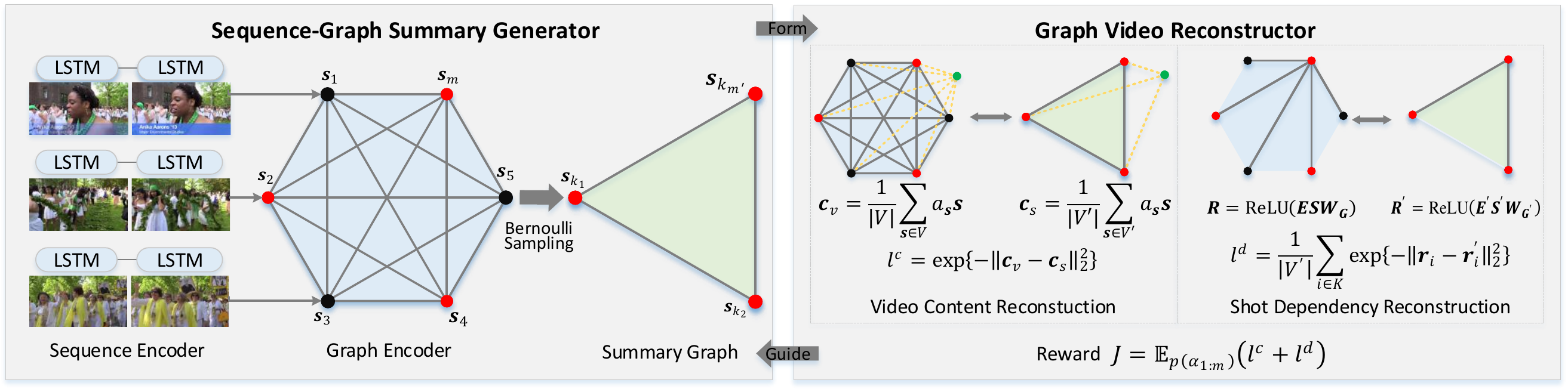}
		\caption{The overview of the proposed Reconstructive Sequence-Graph Network (RSGN). Specifically, the LSTM units are bidirectional. The red and black nodes denote the key and non-key shots, respectively. The two virtual green nodes stand for the encoded video and summary content.}
		\label{fig2}
	\end{figure*}
	
Recently, RNN-based summarization approaches have drawn increasing attention. Zhang \emph{et al.} \cite{DBLP:conf/eccv/ZhangCSG16} firstly utilize a bidirectional LSTM to model the video data as a plain sequence, and predict the probability of each shot to be selected into the summary recursively. Based on the assumption that video data is layered as frames and shots, Zhao \emph{et al.} \cite{DBLP:conf/mm/ZhaoLL17} develop a two-layer LSTM to model the video data hierarchically. Furthermore, the shot segmentation is also considered as an extra task to capture the video structure information \cite{DBLP:conf/cvpr/ZhaoLL18}. The proposed approach also adopts a hierarchical structure to model a video. However, our approach captures the global dependencies among shots using graph convolution network (GCN) rather than LSTM. With the sequence-graph structure, the proposed approach can model a video more comprehensively than RNN-based approaches.
	In general, the above RNN models are trained by simply maximizing the overlap between the predicted summary and human-annotations. To provide more guidance for the training procedure, Mahasseni \emph{et al.} \cite{DBLP:conf/cvpr/MahasseniLT17} present an adversarial LSTM network to generate video summaries discriminatively. Similarly, Fu \emph{et al.} \cite{8658673} propose an attentive and adversarial learning scheme, which can overcome the unbalanced training-test length and leverage the unpaired problems in video summarization. Different from the adversarial loss used in previous works, the proposed reconstruction loss is designed to force the summary generator to preserve the video content and shot-level dependency. Moreover, Zhou \emph{et al.} \cite{DBLP:conf/aaai/ZhouQX18} adopt property models as extra constraints for the generated summaries. However, recent advances have pointed out the deficiency of sequence models in video modeling, where the global dependency is considered via attention mechanism or graph convolutional networks \cite{mao2018hierarchical,chen2019graph,zeng2019graph}. VASNet \cite{fajtl2018summarizing} leverages self-attention mechanism to replace traditional RNN-based modeling to capture the global dependencies. FCSN \cite{rochan2018video} uses 1D fully convolutional network (FCN) to process a frame sequence to capture long-distance relationships. However, VASNet and FCSN both model a video in the frame-level without considering the structure of the video.	Motivated by this, a sequence-graph model is developed in this work.
	
	%------------------------------------------------------------------------
	
	\section{The Proposed Approach}\label{approach}
	\subsection{Overview}
	Focusing on the video summarization task, a Reconstructive Sequence-Graph Network (RSGN) is proposed to 1) address the drawbacks of traditional sequence models in long-distance dependency capturing, and 2) better preserve the video content and shot-level dependencies in the summarization process. As depicted in Fig. \ref{fig2}, it contains a summary generator and a video reconstructor. The generator is in a hierarchical structure, where the first layer is a bidirectional LSTM to encode the frame sequence in each shot, and the second layer is a graph model to explore the dependencies among different shots. After that, the summary candidate is selected and input to the reconstructor.
	
	The reconstructor is in a summary version of the graph encoder. It is utilized to model the summary candidate and measure its reconstructiveness to the original video content and shot-level dependency. In our work, the reconstructor can reward the summary generator while training in a reinforcement scheme. In this case, the summary generator tends to select the key-shots that can better preserve the original video content and shot-level dependency, so that the viewer can understand the main video content from the summary effortlessly. Practically, the proposed approach can be trained either in a supervised or unsupervised manner under the guidance of the reconstructor. Moreover, it should be noted that the reconstructor is only utilized in the training procedure, and the video is summarized purely by the summary generator in the testing procedure.

	\subsection{Sequence-Graph Summary Generator}
	
	\subsubsection{Frame-level Sequence Encoder}
	
	Specifically, given a video with frames \(\left\{ {{\bm{f}_1},{\bm{f}_2}, \cdots ,{\bm{f}_n}} \right\}\), the shot boundaries are firstly detected by Kernel-based Temporal Segmentation (KTS) \cite{potapov2014category}, which is an effective technique in shot boundary detection. In this paper, the indexes  of detected shot boundaries are denoted as \(\left\{ {{b_0},{b_1}, \cdots ,{b_m}} \right\}\), where \({b_0} = 1\) and \({b_m} = n\) stand for the beginning and end of the video.
	
	After the video is segmented, each shot is encoded by the bidirectional LSTM. Taking the \(i\)-th shot as example, the frame subsequence \(\left\{ {{\bm{f}_{{b_{i - 1}}+1}},{\bm{f}_{{b_{i - 1}} + 2}}, \cdots ,{\bm{f}_{{b_i}}}} \right\}\) is encoded as follows,
	\begin{equation}\bm{h}_t^{} = {\rm BiLSTM}\left( {{\bm{f}_t},\bm{h}_{t - 1}^{}} \right),\;t \in \left[ {{b_{i - 1}} + 1,{b_i}} \right],\end{equation}
	where \({\rm BiLSTM}\left(  \cdot  \right)\) is the abbreviation for the calculations in the bidirectional LSTM. It is stacked by two LSTMs, one of which operates reversely. In fact, there are several variants of LSTM, the one proposed in \cite{hochreiter1997long} is widely used in video summarization, which is also adopted in our approach. \(\bm{h}_t\) denotes the hidden state of the bidirectional LSTM. The final hidden state \({\bm{h}_{{b_i}}^{}}\) has encoded both the forward and backward temporal dependencies among frames in the \(i\)-th shot. Naturally, it is taken as the shot feature, \(\bm{s}_i\), and input to the graph model in the second layer.
	
	\subsubsection{Shot-level Graph Encoder}
	
	A video is typically modeled as a frame sequence (Euclidean structure) in previous works \cite{DBLP:conf/eccv/ZhangCSG16,DBLP:conf/mm/ZhaoLL17,DBLP:conf/cvpr/MahasseniLT17}, where the long-term dependencies are difficult to capture because of the inherent limitation of RNN. Recently, Graph Convolutional Network (GCN) \cite{kipf2016semi-supervised} is proposed to model relationships within a non-Euclidean structure such as social network\cite{wu2018socialgcn}, and reveals great potential to exploit internal dependencies in data. In this case, we propose to model a video as a graph, aiming to exploit all the pairwise dependencies between shots. Specifically, the video shots are modeled as a complete graph \(G = \left( {V,E} \right)\). \(V\) is a set of nodes, which is composed of the video shots \(\left\{ {{\bm{s}_1},{\bm{s}_2}, \cdots ,{\bm{s}_m}} \right\}\). \(E\) is a set of edges reflecting the connections between any two nodes. In the complete graph, \(E = \left\{ {{E_{11}}, \cdots ,{E_{ij}}, \cdots ,{E_{mm}}} \right\}\), since all the nodes are connected.
	
	In our approach, each edge $E_{ij}$ in \(E\) is equipped with a weight $e_{ij}$ that indicates the dissimilarity between the connected two shots. We use dissimilarity as the weight because the edges are channels for nodes to aggregate information within the graph, but similar shots do not have to share information since they already have similar semantic features, while shots with huge differences need to communicate for modeling the whole video comprehensively. Inspired by the non-local neural networks \cite{DBLP:conf/cvpr/0004GGH18}, we have tried three functions to compute the edge weights, which are formulated as follows:
	
	\begin{itemize}
		
		\item Dot Product: \begin{equation}{e_{ij}} = -\phi {\left( {{\bm{s}_i}} \right)^{\rm T}}\varphi \left( {{\bm{s}_j}} \right).\end{equation}
		
		\item Gaussian: \begin{equation}{e_{ij}} = \exp \left\{ {-\phi {{\left( {{\bm{s}_i}} \right)}^{\rm T}}\varphi \left( {{\bm{s}_j}} \right)} \right\}.\end{equation}
		
		\item Concatenation: \begin{equation}{e_{ij}} =  {\bm{W}_e^{\rm T}\left[ {\phi \left( {{\bm{s}_i}} \right),\varphi \left( {{\bm{s}_j}} \right)} \right]} .\end{equation}
		
	\end{itemize}
	Specifically, \(i,j = 1,2, \cdots ,m\) for the above three equations. \([ \cdot,\cdot ]\) denotes the concatenation operation. \(\phi \left(  \cdot  \right)\) and \(\varphi \left(  \cdot  \right)\) are the linear embedding functions with \({\bm{W}_\phi }\) and \({\bm{W}_\varphi }\) as the embedding matrices. \({\bm{W}_\phi }\), \({\bm{W}_\varphi }\) and \(\bm{W}_e\) are training parameters.
	
	After the dependencies among video shots are computed, the relationships among all shots in a video is computed by graph convolution \cite{kipf2016semi-supervised}:
	\begin{equation}{\bm{R}} = {\rm ReLU}(\bm{ESW}_{G}),\end{equation}
	where $\bm{E}=(e_{ij})_{m\times m}$, $\bm{S}=\left[\tau\left(\bm{s}_1\right);\tau\left(\bm{s}_2\right);\cdots;\tau\left(\bm{s}_m\right)\right]$ is the embedded feature matrix of all shots ($\tau\left(  \cdot  \right)$ is the linear embedding function parameterized by \(\bm{W}_{\tau}\)), $\bm{W}_G$ is the learnable parameter in graph convolution for graph $G$, \({\rm ReLU}\left(  \cdot  \right)\) is the activation function. $\bm{R}=\left[\bm{r}_1;\bm{r}_2;\cdots;\bm{r}_m \right]$ encodes the relationships among all shots.

	Finally, the probability of being selected as the key-shot is determined jointly by the shot feature and its relationship to the whole video content. It is formulated as
	\begin{equation}{p_i} = {\rm Sigmoid}\left( {{\bm{W}_p}\left[ {\tau\left(\bm{s}_i\right),{\bm{r}_i}} \right] + {b_p}} \right),\end{equation}
	where \({\rm Sigmoid}\left(  \cdot  \right)\) is the activation function, \(\bm{W}_p\) and \(b_p\) are the training weight and bias, respectively. With the predicted probability, the summary candidate is generated by the Bernoulli sampling, \emph{i.e.},
	\begin{equation}\alpha_i = {\rm Bernoulli}\left( {{p_i}} \right),\;i = 1,2, \cdots ,m.\end{equation}
	where \(\alpha_i \in \left\{ {0,1} \right\}\) indicates whether the \(i\)-th shot is selected into the summary or not. 
	
	Besides, when training the summary generator under the supervision of human annotations, the Mean Square Error (MSE) is employed to measure the prediction loss as follows,
	\begin{equation}{l^p} = \frac{1}{m}\left\| {\bm{p} - \bm{g}} \right\|_2^2,\end{equation}
	where \(\bm{p}\) and \(\bm{g}\) denote the score vectors predicted by our approach and annotated by human beings, respectively. \(\left\|  \cdot  \right\|{}_2\) is the \(l_2\) norm.
	
	\subsection{Graph Video Reconstructor}
	
	The reconstructor takes the summary candidate as input, and focuses on the reverse task of video summarization, \emph{i.e.}, reconstructing the original video from the summary. In our work, it is developed as a summary version of the video graph, denoted as \(G'=\left( {V',E'} \right)\). \(V'=\left\{ {{\bm{s}_{k_1}},{\bm{s}_{k_2}}, \cdots ,{\bm{s}_{k_{m'}}}} \right\}\) and \(E'=\left\{ {{E_{k_1,k_1}}, \cdots ,{E_{k_i,k_j}}, \cdots ,{E_{k_{m'},k_{m'}}}} \right\}\) are the key-shot set and their dependencies, where $K=\{k_j\}_{j=1}^{m'}$ is the index set of selected shots. Generally, the target of video reconstructor is to guide the summary generator to select key-shots that contain the main video content and preserve the shot-level dependencies. Therefore, the reconstructor consists of two parts, \emph{i.e.}, the video content reconstruction and shot-level dependency reconstruction.
	\subsubsection{Video Content Reconstruction}
	The summary candidate ought to preserve the main content of the video. To measure the reconstructiveness of the summary to the video content, the video graph \(G\) and its summary version \(G'\) are utilized to model the video content before and after summarization, respectively.
	
	In general, the human-annotated scores of each shot indicate their importance to the video content. It is proper to encode the video content by combining the video shots according to their annotated scores, \emph{i.e.},
	\begin{equation}{\bm{c}_v} = \frac{1}{|V|}\sum\limits_{\bm{s} \in V}{{a_{\bm{s}}}{\bm{s}}}, \end{equation}
	where \(a_{\bm{s}}\) denotes the annotated score of the shot $\bm{s}$. Similarly, the summary content is calculated by
	\begin{equation}{\bm{c}_s} = \frac{1}{|V'|}\sum\limits_{\bm{s} \in V'}{{a_{\bm{s}}}{\bm{s}}}.\end{equation}
	
	After the video and summary content are encoded, the reconstructiveness of the video summary is measured by
	\begin{equation}{l^c} = \exp \left\{ { - \left\| {{\bm{c}_v} - {\bm{c}_s}} \right\|_2^2} \right\}.\end{equation}
	Note that the annotated score \(a_{\bm{s}}\) is not available in the unsupervised setting. To conduct the reconstruction mechanism unsupervisedly, \(a_{\bm{s}}\) in Eqn. (9) and (10) is replaced with the predicted probability in Eqn. (6).
	
	\subsubsection{Shot-level Dependency Reconstruction}
	The shot-level dependencies indicate the storyline of the video. Once the storyline is severely damaged in summarization, the viewer can hardly understand the summary content, let alone infer the original video content. In this case, the shot-level dependencies should be preserved in the summarization task.
	
	In the video, the dependencies of a certain shot \(i\) to others have already been encoded into \(\bm{r}_i\), as formulated in Eqn. (5). Similarly, the dependencies of a key-shot in the summary candidate is encoded as
	\begin{equation}{\bm{R}'} = {\rm ReLU}(\bm{E}'\bm{S}'\bm{W}_{G'}),\end{equation}
	where \(\bm{E}'\) reflects the dependencies among key-shots. As aforementioned, \(G'\) is a summary version of \(G\). Thus, \(\bm{E}'\) is learned similar to \(\bm{E}\) with the trainable parameters \(\left( {{\bm{W}'_\phi },{\bm{W}'_\varphi },{\bm{W}'_e}} \right)\), as formulated in Eqn. (2-4). $\bm{S}'=\left[\tau'\left(\bm{s}_{k_1}\right);\tau'\left(\bm{s}_{k_2}\right);\cdots;\tau'\left(\bm{s}_{k_{m'}}\right)\right]$ is the embedded feature matrix of key-shots ($\tau'\left(  \cdot  \right)$ is the linear embedding function parameterized by \(\bm{W}_{\tau'}\)), $\bm{W}_{G'}$ is the learnable parameter in graph convolution for graph $G'$. 
	
	Note that $G$ and $G’$ have the same purpose so they have the same structure except that the nodes are different. However, since $G$ and $G'$ are constructed from video domain and summary domain, respectively (what they focus on are different), they do not share parameters in consideration of robustness and performance.
	
	Then, the reconstructiveness of the summary to the shot-level dependency is measured by
	\begin{equation}{l^d} = \frac{1}{{\left| {V'} \right|}}\sum\limits_{i\in K} {\exp \left\{ { - \left\| {{\bm{r}_i} - {{\bm{r}'_i}}} \right\|_2^2} \right\}}. \end{equation}
	\subsection{Optimization}
	
	In our work, the summary generator and video reconstructor are optimized in an end-to-end scheme. Particularly, the reinforcement learning \cite{DBLP:journals/ml/Williams92} is adopted to make the reconstructor reward the summary generator effectively.
	
	Practically, the proposed RSGN can be optimized either in a supervised or unsupervised manner. The only difference is whether the prediction loss is adopted for optimization. Take the supervised setting as an example, the proposed approach is optimized under the supervision of the prediction loss (\emph{i.e.}, \(l^p\)) and the reward from the reconstructor (\emph{i.e.}, \(l^c\) and \(l^d\)). Besides, considering that the reconstructor tends to select more key-shots to increase the reward, a regularization term is developed to constrain the size of the summary candidates, which is formulated as
	\begin{equation}{l^r} = {\left( {\frac{1}{m}\sum\limits_{i =1}^{m} {{p_i}}  - \varepsilon } \right)^2},\end{equation}
	where $\varepsilon$ regularizes the distribution of predicted probability to avoid the trivial solution. Following existing approaches \cite{DBLP:conf/cvpr/MahasseniLT17,DBLP:conf/aaai/ZhouQX18}, it is fixed as 0.5 in this paper.
	
	Finally, the loss function is defined as 
	\begin{equation}\mathcal{L} (\bm{\theta}) = {l^p} + {l^r} - J, \end{equation}
	where \(\bm{\theta}\) denotes the training parameter set, and
	\begin{equation}J = {{\mathbb E}_{p\left( {{\alpha _{1:m}}} \right)}}\left( {{l^c} + {l^d}} \right),\end{equation}
	is the expected reward to be maximized, and \(l^p\) is omitted in the unsupervised setting. In our work, the episodic reinforcement algorithm in \cite{DBLP:journals/ml/Williams92} is adopted to optimize Eqn. (15). The derivative of \(J\) can be approximated by \begin{equation}{\nabla _{\bm{\theta}} }J = \frac{1}{n}\sum\limits_{j = 1}^n {\sum\limits_{i = 1}^m {{{\left( {{l^c} + {l^d}} \right)}_j}{\nabla _{\bm{\theta}} }} \log {\pi _{\bm{\theta}} }\left( {{\alpha _i}|{\bm{s}_i},{\bm{r}_i}} \right)}, \end{equation}
	where \({{\pi _{\bm{\theta}} }}\) is the policy function defined by our approach, \(m\) and \(n\) denote the number of shots and episodes. \({\left( {{l^c} + {l^d}} \right)_j}\) is the reward in the \(j\)-th episode. Finally, \({\bm{\theta}}\) is updated by \({\bm{\theta}} ' = {\bm{\theta}}  - \gamma {\nabla _{\bm{\theta}} }\left( {{l^p} + {l^r} - J} \right),\) where \(\gamma\) is the learning rate.
	
	The proposed RSGN is conducted on PyTorch. Specifically, the dimensionality of the hidden states is 256 for both the LSTM and the graph. The full architecture is optimized with the Adam Optimizer, whose learning rate and weight decay rate are both 1e-5, and the learning rate decays every 30 steps with the rate fixed as 0.1. The episode of the reinforcement learning algorithm is set as 10, and the total training epochs are 60.
	
	\section{Experiments}\label{experiment}
	The experiment is conducted on three benchmarks, including SumMe \cite{DBLP:conf/eccv/GygliGRG14}, TVsum \cite{song2015tvsum} and VTW \cite{DBLP:conf/eccv/ZengCNS16}. Several popular supervised and unsupervised approaches are compared to verify the effectiveness of the proposed approach.
	\subsection{Experimental Setup}
	
	\textbf{Datasets.} Five datasets are employed in our experiment. SumMe contains 25 videos with durations varying from 1.5min to 6.5min. Each video is annotated by 15--18 users with both frame-level importance scores and shot-based video summaries. TVsum has 50 videos with durations varying from 2min to 10min. Each video is annotated by 20 users with shot-level importance scores, where the shots are manually generated by segmenting the videos into two-second clips. Following existing protocols, 80\% videos are used for training and the rest 20\% are for testing on both the SumMe and TVsum datasets. Besides, the key-frame based summary datasets, OVP \cite{DBLP:journals/prl/AvilaLLA11} and YouTube \cite{DBLP:journals/prl/AvilaLLA11}, are exploited to augment the training data. They contain 89 videos totally.
	
	VTW is a much larger dataset originally proposed for video captioning. It contains 18100 videos, 2529 of which are annotated with shot-based video highlights. In our work, they are adopted for evaluation by transferring the annotations into frame-level importance scores. Concretely, the highlighted part is scored with 1 and others are 0. 
	
	\textbf{Preprocessing.}
	Similar to existing approaches \cite{DBLP:conf/mm/ZhaoLL17,DBLP:conf/aaai/ZhouQX18,DBLP:conf/eccv/ZhangGS18}, the pool5 layer of GoogLeNet is adopted for frame feature extraction \cite{DBLP:conf/cvpr/SzegedyLJSRAEVR15}, whose dimensionality is 1024. Besides, each video is segmented into shots by KTS \cite{potapov2014category}, which is a widely used video temporal segmentation method in the video summarization task. 
	
	\textbf{Evaluation.}
	The summarization performance is evaluated according to the overlap between the generated summaries and human-created summaries. In our work, F-measure is adopted as the evaluation metric, which is widely used in the summarization task \cite{DBLP:conf/cvpr/ZhaoLL18,li2017general}. Specifically, for both the SumMe and TVsum dataset, each video is annotated with multiple human-created summaries. Following existing protocols, the generated summary is compared with these human-created summaries one-by-one for each video. On SumMe, the maximum evaluation score is taken as the final result. On TVsum, the average of these scores is taken as the final result.
	Furthermore, considering that the summarization results vary largely among videos, the cross-validation is performed in the experiment to make the reported results more convincing. 
	To be specific, the dataset is randomly split into training set (80\%) and testing set (20\%) for 5 independent times. For each split, we train our model on the training set and test it on the corresponding testing set. So, we get 5 results from 5 splits. The final metric is obtained by averaging the 5 results. This strategy is widely adopted in existing approaches \cite{DBLP:conf/eccv/ZhangCSG16,DBLP:conf/mm/ZhaoLL17,DBLP:conf/cvpr/ZhaoLL18,DBLP:conf/aaai/ZhouQX18,DBLP:conf/cvpr/MahasseniLT17}.
	
	In addition, we demonstrate the performance of our approach in terms of rank-based statistics (Kendall’s $\tau$ and Spearman’s $\rho$), which provide additional information on the behavior of summarization approaches \cite{otani2019rethinking}.
	
	\subsection{Results and Discussions}
	\subsubsection{Results on SumMe and TVsum}
	In this part, the ablation studies on several baselines are firstly presented, and then the proposed approach is compared with several state-of-the-arts.
	
	\textbf{Comparison with baselines.} The proposed RSGN is composed of a sequence-graph encoder and a content-dependency reconstructor. To verify the effectiveness of each part, the following baselines are constructed.
	
	\begin{table}[t]
		\centering
		\caption{The comparison of baselines on the SumMe and TVsum datasets. (The scores in bold indicate the best values.)}\label{Table1}
		\renewcommand\arraystretch{1.2}
		
		\begin{tabular}{|p{3cm}<{\centering}||p{1.5cm}<{\centering}|p{1.5cm}<{\centering}|}
			\hline
			Datasets  &SumMe  &TVsum  \\
			\hline
			\hline
			Enc(S), Rec(-)&0.397 &0.556\\
			Enc(G), Rec(-)&0.419 &0.564\\
			Enc(S,S), Rec(-)&0.416 &0.583\\
			Enc(S,G), Rec(-)  &0.428 &0.587 \\
			\hline
			\hline
			Enc(S,G), Rec(C)  &0.436 &0.597 \\
			Enc(S,G), Rec(D)  &0.438 &0.599 \\
			\hline
			\hline
			Enc(G), Rec(D,C)  &0.427 &0.576 \\
			Enc(M,G), Rec(D,C)  &0.415 &0.585 \\		
			\hline
			\hline
			RSGN\(_{uns}\)
			&0.423	&0.580\\
			RSGN&\textbf{0.450}	&\textbf{0.601}\\
			
			\hline
		\end{tabular}
		
	\end{table}
	
	\begin{table}[t]
		\centering
		\caption{The comparison of different edge types of our graph encoder on the SumMe and TVsum datasets.}\label{Table2}
		\renewcommand\arraystretch{1.2}
		
		\begin{tabular}{|p{3cm}<{\centering}||p{1.5cm}<{\centering}|p{1.5cm}<{\centering}|}
			\hline
			Datasets  &SumMe  &TVsum  \\
			\hline
			\hline
			Graph(Dot product)&0.419 &0.564\\
			Graph(Gaussian)	&0.413 &0.558\\
			Graph(Concatenation)&0.415 &0.566\\
			\hline
			
		\end{tabular}
		
	\end{table}

	1) Encoders. Two kinds of encoders are presented, \emph{i.e.}, plain encoders and hierarchical encoders. Plain encoders are denoted as (S) and (G). (S) means the video frames are taken as a sequence and encoded by the bidirectional LSTM. (G) indicates the video data is encoded by our graph model. Hierarchical encoders include (S,S), (S,G). They stand for the hierarchical LSTM and sequence-graph, respectively. To verify the influence of LSTM, we also construct an encoder (M,G) where the feature of each shot is computed by mean pooling of all frame features within the shot. Besides, for the graph encoder, three edge weights are exploited, including dot product, gaussian and concatenation.
	
	2) Reconstructors. Two kinds of reconstructors are developed in our paper, denoted as (C) and (D). (C) stands for the video content reconstructor. (D) represents the shot-level dependency reconstructor. Besides, to verify their effectiveness, the baseline without reconstructor is also tested, denoted as Rec(-). 
	
	Table \ref{Table1} presents the comparisons of different baselines. For the encoders, it can be observed that Enc(G) performs better than Enc(S), which illustrates that it is more proper to model the video data as a graph than a sequence, and has shown the superiority of our graph model in video summarization. In particular, the graph models with different types of edge weight are compared in Table \ref{Table2}. We can see that they get comparable results, so our graph model is robust to the computation functions of edge weight. More specifically, the edge weight computed by the dot product function surpasses the other two slightly. In this case, it is adopted for the following comparison.
	
	Besides, the hierarchical encoders get better results than the plain ones, since the hierarchical structure is more suitable for the two-layered video data and can enhance the nonlinear fitting ability of encoders. Specifically, the performance of Enc(S,G) is better than Enc(S,S). It indicates the effectiveness of the hierarchical encoder with LSTM to capture the frame-level temporal dependency and the graph model to capture the shot-level global dependency.
	\begin{table}[t]
		\centering
		\caption{The comparison of unsupervised approaches on the SumMe and TVsum datasets.}\label{Table3}
		\renewcommand\arraystretch{1.2}
		
		\begin{tabular}{|p{3cm}<{\centering}||p{1.5cm}<{\centering}|p{1.5cm}<{\centering}|}
			\hline
			Datasets  &SumMe  &TVsum  \\
			\hline
			\hline
			Delauny \cite{DBLP:journals/jodl/MundurRY06} &0.315 &0.394\\
			VSUMM \cite{DBLP:journals/prl/AvilaLLA11} &0.335 &0.391\\
			\hline
			\hline
			SALF \cite{DBLP:conf/cvpr/ElhamifarSV12} &0.378 &0.420\\
			LiveLight \cite{DBLP:conf/cvpr/ZhaoX14a} &0.384 &0.477\\
			\hline
			\hline
			SUM-GAN \cite{DBLP:conf/cvpr/MahasseniLT17} &0.387 &0.508\\
			DR-DSN \cite{DBLP:conf/aaai/ZhouQX18} &0.414 &0.576\\
			\hline
			\hline
			RSGN\(_{uns}\) &\textbf{0.423} &\textbf{0.580}\\
			\hline
			
		\end{tabular}
		
	\end{table}
	
	\begin{table}[t]
		\centering
		\caption{The comparison of supervised approaches on the SumMe and TVsum datasets.}\label{Table4}
		\renewcommand\arraystretch{1.2}
		
		\begin{tabular}{|p{3cm}<{\centering}||p{1.5cm}<{\centering}|p{1.5cm}<{\centering}|}
			\hline
			Datasets  &SumMe  &TVsum  \\
			\hline
			\hline
			CSUV \cite{DBLP:conf/eccv/GygliGRG14} &0.393 &0.532\\
			LSMO \cite{gygli2015video} &0.403 &0.568\\
			\hline
			\hline
			vsLSTM \cite{DBLP:conf/eccv/ZhangCSG16} &0.376 &0.542\\
			dppLSTM \cite{DBLP:conf/eccv/ZhangCSG16} &0.386 &0.547\\
			SUM-GAN\(_{sup}\) \cite{DBLP:conf/cvpr/MahasseniLT17} &0.417 &0.563\\
			DR-DSN\(_{sup}\) \cite{DBLP:conf/aaai/ZhouQX18} &0.425 &0.581\\
			
			H-RNN \cite{DBLP:conf/mm/ZhaoLL17} &0.421 &0.579\\
			HSA-RNN \cite{DBLP:conf/cvpr/ZhaoLL18} &0.423 &0.587\\
			re-SEQ2SEQ \cite{DBLP:conf/eccv/ZhangGS18} &0.425 &\textbf{0.603}\\
			VASNet \cite{fajtl2018summarizing} &0.424&0.589\\
			\hline
			\hline
			RSGN &\textbf{0.450} &0.601\\
			\hline
			
		\end{tabular}
		
	\end{table}
	
	\begin{table*}[t]
		\centering
		\caption{The results with different training settings on the SumMe and TVsum datasets.}\label{Table5}
		\renewcommand\arraystretch{1.2}	
		\begin{tabular}{|p{3cm}<{\centering}||p{1.6cm}<{\centering}|p{1.6cm}<{\centering}|p{1.6cm}<{\centering}||p{1.6cm}<{\centering}|p{1.6cm}<{\centering}|p{1.6cm}<{\centering}|}
			\hline
			Datasets &\multicolumn{3}{c||}{SumMe}&\multicolumn{3}{c|}{TVsum}\\
			\hline	
			Approaches  &Canonical  &Augmented  &Transfer&Canonical  &Augmented  &Transfer\\
			\hline
			\hline	
			SUM-GAN \cite{DBLP:conf/cvpr/MahasseniLT17} &0.387 &0.417&--&0.508 &0.589 &--\\
			DR-DSN \cite{DBLP:conf/aaai/ZhouQX18} &0.414 &0.428&0.424&0.576 & 0.584 &0.578  \\
			\hline
			\hline
			vsLSTM \cite{DBLP:conf/eccv/ZhangCSG16} &0.376 &0.416&0.407&0.542 & 0.579 &0.569  \\
			dppLSTM \cite{DBLP:conf/eccv/ZhangCSG16} &0.386 &0.429&0.418&0.547 & 0.596 &0.587  \\
			SUM-GAN\(_{sup}\) \cite{DBLP:conf/cvpr/MahasseniLT17} &0.417 &0.436&--&0.563 &0.612 &--\\
			DR-DSN\(_{sup}\) \cite{DBLP:conf/aaai/ZhouQX18} &0.421 &0.439&0.426&0.581 & 0.598 &0.589  \\
			H-RNN \cite{DBLP:conf/mm/ZhaoLL17} &0.421 &0.438&--&0.579 & 0.619 &--  \\
			HSA-RNN \cite{DBLP:conf/cvpr/ZhaoLL18} &0.423 &0.421&--&0.587 & 0.598 &--  \\
			re-SEQ2SEQ \cite{DBLP:conf/eccv/ZhangGS18} &0.425 &0.449&--&\textbf{0.603} &\textbf{ 0.639} &--  \\
			VASNet \cite{fajtl2018summarizing} &0.424 &0.425&0.419&0.589 & 0.585 &0.547  \\
			\hline
			\hline
			RSGN\(_{uns}\) &0.423	&0.436	&0.412	&0.580	&0.591	&0.597\\
			RSGN &\textbf{0.450}	&\textbf{0.457}	&\textbf{0.440}&0.601	&0.611	&\textbf{0.600}\\
			\hline	
		\end{tabular}	
	\end{table*}
	
	\begin{figure*}
		\centering
		\includegraphics[width=0.98\textwidth]{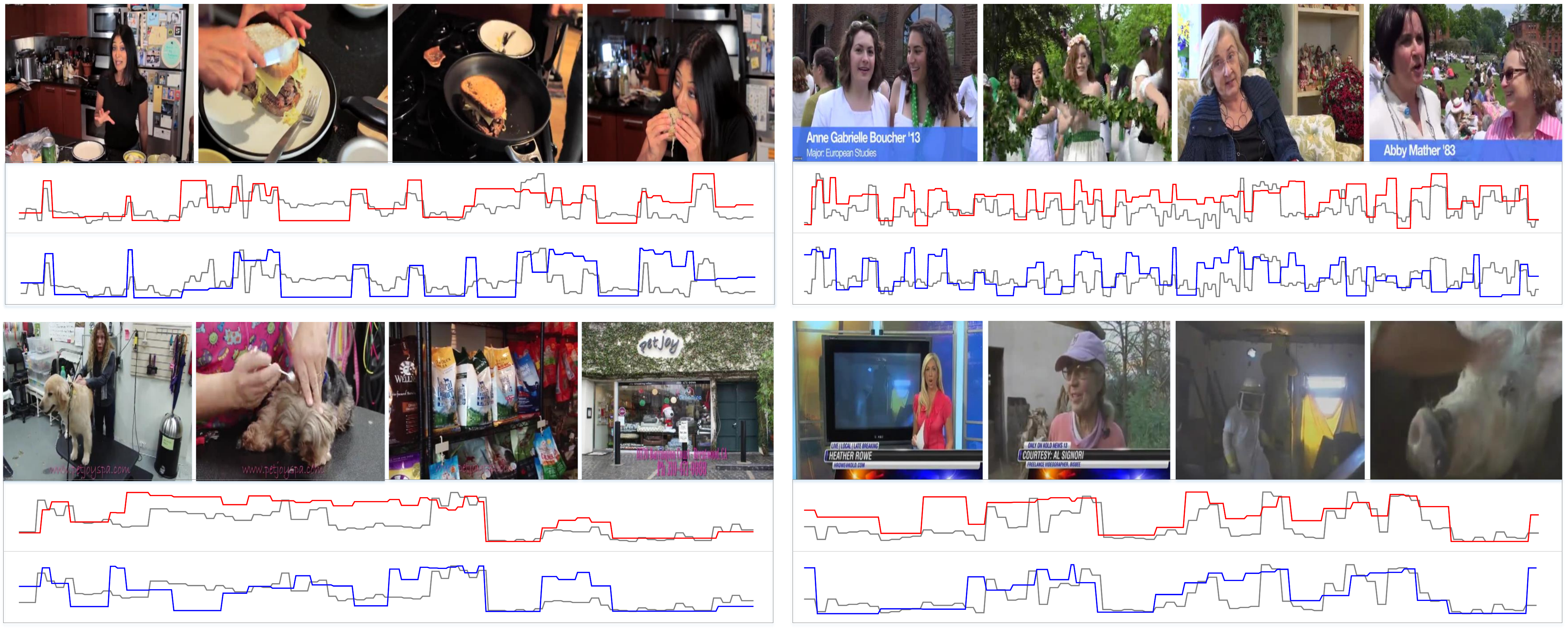}
		\caption{The summarization results of RSGN and RSGN\(_{uns}\). The images are sampled from the summaries generated by RSGN. The curves denote the distributions of importance scores. The gray curves depict the ground truth score, while the red/blue curves depict the score predicted by the supervised/unsupervised model, respectively.}
		\label{fig3}
	\end{figure*}
	
	For reconstructors, Rec(-) denotes the proposed approach trained by simply maximizing the overlap between the human-created summaries and automatically generated summaries. It can be seen that Rec(-) shows less effectiveness than the others equipped with reconstructors, including Rec(C), Rec(D) and Rec(C,D). It exhibits the improvements of the reconstructor to the summary quality. Although no extra supervision is required, it can indeed provide more guidance for the summary generation process. Specifically, Rec(C) and Rec(D) perform comparably, and our full approach RSGN with Rec(C,D) outperforms them. It demonstrates both the video content and dependencies among shots should be preserved explicitly while summarizing the videos. This point has also been proved by the results of our unsupervised variant RSGN\(_{uns}\), since it performs even better than those supervised baselines without the video reconstructor. 
	
	Compared to the full model RSGN, the performance of RSGN (Enc(G), Rec(D,C)) drops as expected. Since the graph in our model tends to capture global dependencies in the video, abandoning LSTM means the local information is omitted. To better model the global dependencies as well as the local ones, sequence-graph model is crucial in our approach. By comparing RSGN (Enc(M,G), Rec(D,C)) and RSGN, we conclude that sequence modeling by LSTM is crucial for video summarization.
	
	\textbf{Comparison with unsupervised approaches.} Table \ref{Table3} shows the results of various unsupervised approaches on SumMe and TVsum. The compared approaches are in three kinds. Specifically, Delauny and VSUMM summarize the video by delauny clustering and k-means, respectively. They aggregate the video shots depending on their local similarity and select the cluster centers as the key shots. SALF and LiveLight are based on dictionary learning, which summarize the video by measuring the representativeness of each key-shot to the video content. They take the global dependency among video shots into consideration, and get better performance than clustering-based approaches. It shows the necessity of the global dependency in video summarization.
	
	SUM-GAN and DR-DSN are RNN-based approaches. Benefiting from the great ability of RNN in sequence modeling, they perform much better than clustering and dictionary learning based approaches. Specifically, SUM-GAN develops a discriminator to guarantee the generated summary can reconstruct the original video content, which verifies the necessity of reconstructiveness in video summarization. DR-DSN summarizes the video under the guidance of summary properties. The better performance of our approach is mainly owing to our sequence-graph model in frame-level and shot-level dependency capturing, and the graph reconstructor in rewarding the summary generator.
	
	\textbf{Comparison with supervised approaches.}
	Table \ref{Table4} presents the results of supervised approaches on the SumMe and TVsum datasets. CSUV and LSMO summarize the video by designing several property-models.  Other than them, all the compared approaches are based on RNN, which have achieved state-of-the-arts. Specifically, vsLSTM, dppLSTM, SUM-GAN\(_{sup}\), DR-DSN\(_{sup}\) and VASNet encode the video as a long sequence with the plain LSTM. vsLSTM is optimized by simply maximizing the overlap between the automatically predicted summary and the reference. However, it is insufficient to provide the summary generator with enough guidance. In this case, dppLSTM, SUM-GAN\(_{sup}\) and DR-DSN take the summary properties into consideration to boost the performance. In addition, VASNet leverages self-attention mechanism to replace traditional RNN-based modeling for the efficiency of training and inference. The even better performance of our RSGN shows the superiority of our sequence-graph encoder compared to the plain sequence encoder.
	
	Actually, the video data are layered as frames and shots. To adapt the characteristics of video data, hierarchical structures of LSTM are developed in re-SEQ2SEQ, H-RNN and HSA-RNN. They get better results than the plain LSTMs, which has verified the effectiveness of the hierarchical structure. Actually, the proposed RSGN is also in a hierarchical structure. Compared to the other three, the main difference lies in that the second layer of RSGN is a graph model. In this case, the better performance of our RSGN indicates that it is more suitable to model the shot-level dependency as a graph, in order to exploit the global dependency for key-shot selection.
	
	\begin{table*}[t]
		\centering
		\caption{Rank-based evaluation results on SumMe and TVsum.}\label{Table7}
		\renewcommand\arraystretch{1.2}
		
		\begin{tabular}{|p{4cm}<{\centering}||p{1.6cm}<{\centering}|p{1.6cm}<{\centering}||p{1.6cm}<{\centering}|p{1.6cm}<{\centering}|}
			\hline
			Datasets &\multicolumn{2}{c||}{SumMe}&\multicolumn{2}{c|}{TVsum}\\
			\hline	
			Rank Correlation Coefficient   &Kendall’s $\tau$  &Spearman’s $\rho$  &Kendall’s $\tau$  &Spearman’s $\rho$ \\
			\hline
			\hline	
			Random &0.000&0.000&0.000&0.000 \\
			Human &0.205&0.213&0.177&0.204 \\
			\hline
			\hline
			DR-DSN \cite{DBLP:conf/aaai/ZhouQX18} & 0.047 & 0.048 & 0.020 & 0.026 \\
			dppLSTM \cite{DBLP:conf/eccv/ZhangCSG16} &--&--&0.042&0.055 \\
			HSA-RNN \cite{DBLP:conf/cvpr/ZhaoLL18} &0.064&0.066&0.082&0.088\\
			\hline
			\hline
			RSGN\(_{uns}\)	&0.071	&0.073	&0.048	&0.052\\
			RSGN 	&\textbf{0.083}	&\textbf{0.085}	&\textbf{0.083}&\textbf{0.090}\\
			\hline	
		\end{tabular}
		
	\end{table*}
	
	Table \ref{Table5} shows the results of approaches optimized in different training data organizations, including canonical, augmented and transfer, similar to the settings in \cite{DBLP:conf/eccv/ZhangCSG16,DBLP:conf/aaai/ZhouQX18,DBLP:conf/cvpr/MahasseniLT17}. Specifically, the results of unsupervised and supervised approaches are both presented. We can see that the augmented and transfer settings get better results than the canonical setting for most approaches, since more annotated data are employed for training. Particularly, we can see that RSGN\(_{uns}\) performs comparable with those supervised ones in most occasions. It mainly benefits from our sequence-graph encoder and the content-dependency reconstructor.
	
	Figure \ref{fig3} displays some summary examples generated by the proposed RSGN. From the displayed frames, we can easily imagine the activities taking place in the video, which indicates the video content is effectively summarized. Besides, the predicted frame-level scores are visualized under the summaries. We can see that the predicted scores of RSGN and RSGN\(_{uns}\) are quite similar with each other, and both of them can fit the ground truth well. It illustrates that the reward from our reconstructor can imitate the supervision of human annotations, and provide sufficient guidance for learning the latent summarization patterns. It is also the explanation to the comparable performance between RSGN\(_{uns}\) and some supervised approaches.
	\subsection{Results on VTW}
	\begin{table}[t]
		\centering
		\caption{The summarization results on the VTW dataset.}\label{Table6}
		\renewcommand\arraystretch{1.2}
		
		\begin{tabular}{|p{4cm}<{\centering}||p{1.5cm}<{\centering}|}
			\hline
			Metric &F-score\\
			\hline	
			\hline
			vsLSTM \cite{DBLP:conf/eccv/ZhangCSG16} &0.441\\
			dppLSTM \cite{DBLP:conf/eccv/ZhangCSG16} &0.443\\
			H-RNN \cite{DBLP:conf/mm/ZhaoLL17} &0.465\\
			HSA-RNN \cite{DBLP:conf/cvpr/ZhaoLL18} &0.491\\
			re-SEQ2SEQ \cite{DBLP:conf/eccv/ZhangGS18} &0.480\\
			VASNet \cite{fajtl2018summarizing}&0.478\\
			\hline
			\hline
			RSGN (Enc(G), Rec(-)) &0.450\\
			RSGN (Enc(G), Rec(C,D)) &0.468\\
			RSGN\(_{uns}\) &0.471\\
			RSGN &\textbf{0.496}\\
			\hline
			
		\end{tabular}
		
	\end{table}

	Table \ref{Table6} presents the results on the VTW dataset. Similar to the results on SumMe and TVsum, the hierarchical encoders get better results than the plain ones on VTW, not only for the pure RNN-based approaches, but also for the proposed approach, since our RSGN surpasses the two variants with a simple graph encoder, including RSGN (Enc(G), Rec(-)) and RSGN (Enc(G), Rec(C,D)). More specifically, RSGN (Enc(G), Rec(-)) outperforms vsLSTM and dppLSTM that tackle the video data as a plain sequence. It indicates the effectiveness of the graph model in video data encoding. The even better performance of our full model RSGN also demonstrates that it is more proper to encode the video data hierarchically by the sequence-graph model. Specifically, the frames are viewed as sequences and encoded with a bidirectional LSTM, and the shots are encoded by a graph to capture the dependencies among them.
	
	From Table \ref{Table6}, we can clearly see that RSGN (Enc(G), Rec(C,D)) outperforms RSGN (Enc(G), Rec(-)). More particularly, the unsupervised variant of our approach, RSGN\(_{uns}\), gets comparable results with most of the compared supervised approaches. The two points mainly benefit from our reconstructor. By training in a reinforcement scheme, the reconstructor can reward the summary generator in order to provide sufficient guidance for the summarization process. Under this circumstance, the summary generator can be optimized effectively even without the supervision of human-created summaries.
	
	\subsection{Rank-based Evaluation}
	
	The evaluation of video summary is a quite complex task. Well-used F-measure is determined by simply computing the temporal overlap between the generated summary and human annotations, which cannot reflect the summary quality comprehensively. Recently, researchers \cite{otani2019rethinking} have revealed that the distribution of the segment lengths has great impact on the metric and sometimes randomly generated summary achieves considerable F-measure, which means F-measure only is inadequate to reflect the performance of video summarization approaches. In this case, rank-based evaluation for video summarization is proposed \cite{otani2019rethinking}, \emph{i.e.,} evaluating the predicted important scores with respect to the ground truth scores using rank correlation coefficients such as Kendall’s $\tau$ and Spearman’s $\rho$. Our approaches are compared with several importance score based approaches (as well as the randomly generated summaries and human annotations) under the same experiment settings. The comparison results are demonstrated in Table \ref{Table7}. Note that the results of human summary are obtained using leave-one-out approach as in \cite{otani2019rethinking}.
	
	As we can see from Table \ref{Table7}, randomly generated summaries show no correlation with the ground truth annotations, while human annotations reveal highest correspondences with each other, which indicates the internal consistency within human summaries. In terms of the performance of compared approaches, the supervised version of RSGN achieves the best performance in both Kendall’s $\tau$ and Spearman’s $\rho$ on SumMe and TVsum, which means it captures the importance statistics of each video more accurately than other approaches. Note that the unsupervised version of RSGN also achieves comparable performance. To be specific, the results of our unsupervised version are much higher than those of unsupervised approaches, DR-DSN, and higher than those of supervised approaches, such as dppLSTM \cite{DBLP:conf/eccv/ZhangCSG16} and HSA-RNN \cite{DBLP:conf/cvpr/ZhaoLL18}. In this case, our approach is capable of extract informative characteristic within each video initiatively by modeling the reconstruction ability of generated summaries with or without the guidance of human annotations.
	
	\section{Conclusions} \label{conclusion}
	In this paper, a Reconstructive Sequence-Graph Network (RSGN) is proposed for the video summarization task. It targets in addressing the deficiency of pure sequence models in global dependency capturing, and better preserving the original video content and shot-level dependencies when summarizing the video. Specifically, RSGN contains a summary generator and a video reconstructor. The summary generator encodes the video frames and shots as short sequences and a complete graph hierarchically, in order to jointly exploit the frame-level temporal dependencies and shot-level global dependencies. The reconstructor is in a summary version of the graph encoder. It measures the reconstructiveness of the summary to the original video content and shot-level dependencies. In the training procedure, the reconstructor can reward the summary generator so that more guidance are provided. Practically, the results on three benchmarks have verified the effectiveness of the sequence-graph summary generator and the graph reconstructor.
	\bibliographystyle{IEEE}
	\bibliography{bare_adv}

	% biography section
	% 
	% If you have an EPS/PDF photo (graphicx package needed) extra braces are
	% needed around the contents of the optional argument to biography to prevent
	% the LaTeX parser from getting confused when it sees the complicated
	% \includegraphics command within an optional argument. (You could create
	% your own custom macro containing the \includegraphics command to make things
	% simpler here.)
	%\begin{IEEEbiography}[{\includegraphics[width=1in,height=1.25in,clip,keepaspectratio]{mshell}}]{Michael Shell}
	% or if you just want to reserve a space for a photo:

	% if you will not have a photo at all:
	%\begin{IEEEbiographynophoto}{John Doe}
	%Biography text here.
	%\end{IEEEbiographynophoto}
	%
	%% insert where needed to balance the two columns on the last page with
	%% biographies
	%%\newpage
	%
	%\begin{IEEEbiographynophoto}{Jane Doe}
	%Biography text here.
	%\end{IEEEbiographynophoto}
	
	% You can push biographies down or up by placing
	% a \vfill before or after them. The appropriate
	% use of \vfill depends on what kind of text is
	% on the last page and whether or not the columns
	% are being equalized.
	
	%\vfill
	
	% Can be used to pull up biographies so that the bottom of the last one
	% is flush with the other column.
	%\enlargethispage{-5in}

	% that's all folks
\end{document}